\documentclass[11pt]{article}      \usepackage{graphicx}
\usepackage[utf8]{inputenc}
\usepackage{caption}
\usepackage{amssymb}
\usepackage{epsfig}
\usepackage[hyphens]{url}
\usepackage{authblk}
\usepackage{graphicx}
\usepackage{subcaption}
\begin{document}

\title{Interacting Behavior and Emerging Complexity\thanks{Corresponding authors:
hector.zenil at algorithmicnaturelab.org, jjoosten at ub.edu}}

\author{Alyssa Adams$^{\dagger,\top}$, Hector Zenil$^{\ddagger,\top}$, Eduardo Hermo Reyes$^\mp$, Joost J. Joosten$^{\mp,\top}$\\
$^\dagger$Beyond Center, Arizona State University, Tempe, AZ, U.S.A.\\
$^\ddagger$Department of Computer Science, University of Oxford, U.K.\\
$^\mp$Department of Logic, History and Philosophy of Science, University of Barcelona, Spain\\
$^\top$Algorithmic Nature Group, LABoRES, Paris, France
}

\date{}

\maketitle

\begin{abstract}
Can we quantify the change of complexity throughout evolutionary processes? We attempt to address this question through an empirical approach. In very general terms, we simulate two simple organisms on a computer that compete over limited available resources. We implement Global Rules that determine the interaction between two Elementary Cellular Automata on the same grid. Global Rules change the complexity of the state evolution output which suggests that some complexity is intrinsic to the interaction rules themselves. The largest increases in complexity occurred when the interacting elementary rules had very little complexity, suggesting that they are able to accept complexity through interaction only. We also found that some Class 3 or 4 CA rules are more fragile than others to Global Rules, while others are more robust, hence suggesting some intrinsic properties of the rules independent of the Global Rule choice. We provide statistical mappings of Elementary Cellular Automata exposed to Global Rules and different initial conditions onto different complexity classes.\\

\end{abstract}

\section{Introduction}
\label{intro}
The race to understanding and explaining the emergence of complex structures and behavior has a wide variety of participants in contemporary science, whether it be sociology, physics, biology or any of the other major sciences. One of the more well-trodden paths is one where evolutionary processes play an important role in the emergence of complex structures. When various organisms compete over limited resources, complex behavior can be beneficial to outperform competitors. But the question remains: Can we \emph{quantify} the change of complexity throughout evolutionary processes? The experiment we undertake in this paper addresses this question through an empirical approach. In very general terms, we simulate two simple organisms on a computer that compete over limited available resources. 
% XXX include two GNS quotes here; Bennett, etc.

Since this experiment takes place in a rather abstract modeling setting, we will use the term \emph{processes} instead of organisms from now on. Two competing processes evolve over time and we measure how the complexity of the emerging patterns evolves as the two processes interact. The complexity of the emerging structures will be compared to the complexity of the respective processes as they would have evolved without any interaction.

When setting up an experiment, especially one of an abstract nature, a subtle yet essential question emerges. \emph{What} exactly defines a particular process and how does the process distinguish itself from the background? For example, is the shell of a hermit crab part of the hermit crab even though the shell itself did now grow from it? Do essential bacteria that reside in larger organisms form part of that larger organism? Where is the line drawn between an organism and its environment? Throughout this paper, we assume that a process is only separate from the background in a behavioral sense. In fact, we assume processes are different from backgrounds in a physical sense only through resource management. We will demonstrate how these ontological questions are naturally prompted in our very abstract modeling environment.

\section{Methods}
\label{sec:1}

As mentioned already, we wish to investigate how complexity evolves over time when two processes compete for limited resources. We chose to represent competing processes by cellular automata as defined in the subsection below. Cellular automata (CA/CAs) represent a simple parallel computational paradigm that admit various analogies to processes in nature. In this section, we shall first introduce Elementary Cellular Automata to justify how they are useful for modeling interacting and competing processes.

\subsection{Elementary Cellular Automata}

We will consider \emph{Elementary Cellular Automata} (ECA/ECAs) which are simple and well-studied CAs. They are defined to act on a one-dimensional discrete space divided into countably infinite many cells. We represent this space by a copy of the integers $\mathbb Z$ and refer to it as a \emph{tape} or simply \emph{row}. 

Each cell in the tape can have a particular color. In the case of our ECAs we shall work with just two colors, represented for example by $0$ and $1$ respectively. We sometimes call the distributions of $0$s and $1$s over the tape the \emph{output space} or \emph{spatial configuration}. Thus, we can represent an output space by a function 
\[
r: \mathbb Z \to \{ 0,1\} \ \ \ \ \ \mbox{(also written $r\in \{ 0,1\} ^\mathbb Z$)}
\]
from the integers $\mathbb Z$ to the set of colors, in this case $\{ 0,1\}$. Instead of writing $r(i)$, we shall often write $r_i$ and call $r_i$ the color of the $i$-th cell of row $r$.

An ECA $\epsilon$ will act on such an output space in discrete time steps and as such can be seen as a function mapping functions to functions. In our case:
\[
\epsilon : \{ 0,1\} ^\mathbb Z \to \{ 0,1\} ^\mathbb Z.
\]
If some ECA $\epsilon$ acts on some initial row $r^0$ we will denote the output space after $t$ time-steps by $r^t$ so that $r^{t+1} := \epsilon (r^t)$. Likewise, we will denote the $n$th cell at time $t$ by $r^t_n$. Our ECAs are entirely defined in a local fashion in that the color $r^{t+1}_n$ of cell $n$ at time $t+1$ will only depend on $r^t_n$ and its two direct neighboring cells $r^t_{n-1}$ and $r^t_{n-1}$ (in more general terminology, we only consider radius-one CAs (\cite{wolfram})). 

Thus, an ECA $\epsilon$ with just two colors in the output space is entirely determined by its behavior on three adjacent cells. Since each cell can only have two colors, there are $2^3=8$ such possible triplets so that there are $2^8=256$ possible different ECAs. However, we will not consider all ECAs for this experiment for computational simplicity. Instead, we only consider the 88 ECA that are non-equivalent under horizontal translation, 1 and 0 exchanges, or any combination of the two. 

\subsection{Interacting and competing ECAs: global rules}

In our experiment, the entire interaction will be modeled by a particular CA that is a combination of two ECAs and something else. So far we have decided to model a process $\Pi$ by an ECA $\epsilon$ with a  color 0 (white) and a non-white color 1. A process is modeled by the evolution of the white and non-white cells throughout time steps as governed by the particular ECA rule.

Once we have made this choice, it is natural to consider a different process $\Pi'$ in a similar fashion. That is, we model $\Pi'$ also by an ECA $\epsilon'$. To tell $\epsilon$ and $\epsilon'$ apart on the same grid we choose $\epsilon'$ to work on the alphabet $\{0,2\}$ whereas the alphabet of $\epsilon$ was $\{0,1\}$. Both $\epsilon$ and $\epsilon'$ will use the same symbol 0 (white).

The next question is how to model the interaction between $\epsilon$ and $\epsilon'$. We will do so by embedding both CAs in a setting where we have a global CA $E$ with three colors $\{ 0,1,2 \}$. We will choose $E$ in such a fashion that $E$ restricted to the alphabet $\{ 0,1\}$ is just $\epsilon$ while $E$ restricted to the alphabet $\{ 0,2\}$ is $\epsilon'$.

Of course these two requirements do not determine $E$ for given $\epsilon$ and $\epsilon'$. Thus, this leaves us with a difficult modeling choice reminiscent to the ontological question for organisms: what to do for triplets that contain all three colors or are otherwise not specified by $\epsilon$ or $\epsilon'$. Since there are 12 such triplets\footnote{These 12 different triplets are $\{ 0, 1, 2\}$, $\{ 0, 2, 1\}$, $\{ 1, 0, 2\}$, $\{ 2, 0, 1\}$, $\{ 1, 2, 0\}$, $\{ 2, 1, 0\}$, $\{ 1, 1, 2\}$, $\{ 1, 2, 1\}$, $\{ 2, 1, 1\}$, $\{ 1, 2, 2\}$, $\{ 2, 1, 2\}$, and $\{ 2, 2, 1\}$.} we have $3^{12}=531\,441$ different ways to define $E$ given $\epsilon$ and $\epsilon'$. Given ECAs $\epsilon$ and $\epsilon'$ as above, we call any such $E$ that extends $\epsilon$ and $\epsilon'$ a corresponding \emph{global rule}. Since there are 88 unique ECAs, there are 3916 unique combinations of $\epsilon$ and $\epsilon'$, which results in $3916 \times 531\,441$ possible globals rules. An online program illustrating interacting cellular automata via a global rule can be visited at \url{http://demonstrations.wolfram.com/CompetingCellularAutomata/}.

\subsection{Intrinsic versus evolutionary complexity}

Let us go back to the main purpose of the paper. We wish to study two competing processes that evolve over time and we want to measure how the complexity of the emerging patterns evolves as the two processes interact. The complexity of the emerging structures will be compared to the complexity of the respective processes as they would have evolved without interacting with each other. 

The structure of an experiment readily suggests itself. Pick two ECAs $\epsilon$ and $\epsilon'$ defined on the alphabets $\{0,1\}$ and $\{0,2\}$ respectively. Measure the typical complexities $c$ and $c'$ generated by $\epsilon$ and $\epsilon'$ respectively when they are applied in an isolated setting. Next, pick some corresponding global rule $E$ and measure the typical complexity $\tilde c$ that is generated by $E$.

Once this `typical complexity' is well-defined, the experiment can be run and $\tilde c$ can be compared to $c$ and $c'$. The question is how to interpret the results. If we see a change in the typical complexity there are three possible reasons this change can be attributed to:
\begin{enumerate}
\item
An evolutionary process triggered by the interaction between $\epsilon$ and $\epsilon'$;

\item
An intrinsic complexity injection due the nature of how $E$ is defined on the 12 previously non-determined tuples;

\item
An intrinsic complexity injection due to scaling the alphabet from size 2 to size 3.

\end{enumerate}

In case of the second reason, an analogy to the cosmological constant is readily suggested. Recall that $E$ is supposed to take care of the modeling of the background so to say, where none of $\epsilon$ or $\epsilon'$ is defined but only their interaction. Thus, a possible increase of complexity/entropy is attributed in the second reason to some intrinsic entropy density of the background. We shall see how the choice of $E$ will affect the change of complexity upon interaction. Before we can further describe the experiment, we first need to decide on how to measure complexity.

\subsection{Characterizing complexity}\label{section:complexityCharacterization}

In \cite{wolfram} a qualitative characterization of complexity is given for dynamical processes in general and ECAs in particular. More specifically, Wolfram described four  classes for complexity which can be characterized as follows:

\begin{itemize}
\item {Class 1.} Symbolic systems which rapidly converge to a uniform state. Examples are ECA rules\footnote{Throughout this paper we will use the Wolfram enumeration of ECA rules, see \cite{wolfram}.} 0, 32 and 160.
\item {Class 2.} Symbolic systems which rapidly converge to a repetitive or stable state. Examples are ECA rules 4, 108 and 218.
\item {Class 3.} Symbolic systems which appear to remain in a random state. Examples are ECA rules 22, 30, 126 and 189.
\item {Class 4.} Symbolic systems which form areas of repetitive or stable states, but which also form structures that interact with each other in complicated ways. Examples are ECA rules 54 and 110.
\end{itemize}

It turns out that one can consider a quantitative complexity measure that largely generalizes the four qualitative complexity classes as given by Wolfram. In our definition of doing so we shall make use of De Bruijn sequences $B(k,n)$ although there are different approaches possible. 

A \emph{de Bruijn} sequence for n colors $B(k,n)$ is any sequence $\sigma$ of n colors so that any of the $n^k$ possible string of length $k$ with $n$ colors is actually a substring of $\sigma$. In this sense, de Bruijn sequences are sometimes considered to be semi-random. Using de Bruijn sequences of increasing length for a fixed number of colors we can parametrize our input and as such it makes sense to speak of asymptotic behavior.

Now, one can formally characterize Wolfram's classes in terms of Kolmogorov complexity as was done in~\cite{zenilchaos}. For large enough input $i$ and for a sufficiently long time evaluation $t$ one can use Kolmogorov complexity to assign a complexity measure $W(\epsilon(i,t))$ of a system  $\epsilon$ for input $i$ and runtime $t$ (see \cite{zenilchaos}).  A typical complexity measure for $\epsilon$ can then be defined by:

\[
W(\epsilon)=\limsup_{i,t\rightarrow\infty}{W(\epsilon(i,t))}.
\]

As a first approximation, one can use cut-off values of these $W(\epsilon)$ outcomes to classify $\epsilon$ into one of the four Wolfram classes.

\subsection{Experiment setup}
In the experiment we tested the influence of global rules (GRs) $E$ on two interacting ECAs $\epsilon$ and $\epsilon'$. Of the $3^{12} = 531\,441$ possible GRs we explored a total number of $425\,600$ GRs representing a share\footnote{We aimed at simulating all GRs but due to some technical restrictions only 80\% of them got executed. For as far as we know, no bias has been introduced by this.} of 80\%. 

For each combination of the two ECAs $\epsilon$ and $\epsilon'$ with one GR, we considered 100 different initial conditions of length 26 corresponding to the first 100 (in the canonical enumeration order) de Bruijn sequences of this length over the alphabet $\{ 0,1,2 \}$. Each execution was evaluated for 60 timesteps. We use the method described in Section \ref{section:complexityCharacterization} to determine the complexity of a CA state evolution for two ECAs with mixed neighborhoods evolving under a GR. Specifically, we used the \textit{Mathematica} v. 10.0 \texttt{Compress} function as an approximation of the Kolmogorov complexity~\footnote{An online program showing how compression characterizes cellular automata evolution can be found in \url{http://demonstrations.wolfram.com/CellularAutomatonCompressibility/}.} as first suggested in~\cite{zenilca}. 

It is important to note the number of computational hours it took to undergo this experiment. For $425\,600$ GRs each exploring 100 different initial conditions and $3\,916$ combinations of $\epsilon$ and $\epsilon'$, there were a total of $166$ billion CA executions. Even with one numerical value per CA execution, the total data generated for this experiment was 1.3 terabytes on hard drive. About $100\,000$ hours of computational time were used total on about 200 jobs at a time, each GR on a different core, which was 200 GRs. Sometimes up to 500 jobs in parallel. Each job took about 36 hours to complete.

Each complexity estimate $c$ is normalized by subtracting the compression value of a string of 0's of equal length. For most of the CA instances, this is a good approximation of the long-term effects of a GR. As mentioned in Section \ref{section:complexityCharacterization} we can use these compression values to determine/approximate Wolfram's complexity classes according to critical thresholds values. The thresholds were trained according to the ECAs we used for the interactions.

Using these compression methods for each run, we determined if the output was Class 1, 2, 3 or 4. The outcome is organized according to the complexity classes of the constituent CAs. For the best clarity, the results are represented in heat maps in the next section.

\section{Results}

Each output class (1-4) is represented by its own heat map, so each figure will have four heat maps to account for the different classes of outputs. Each map is composed of a four-by-four grid whose axes describe the complexity classes of $\epsilon$ and $\epsilon'$, as shown. Thus, of all of the runs (ECAs $\epsilon$ and $\epsilon'$ interaction, GR $E$, and initial condition) that yield a Class 1 output are represented under the heat map labeled \textit{Class 1}. For example, in Figure 1, about 11\% of all Class 1 outputs were generated by Class 1 $\epsilon$ and Class 1 $\epsilon'$ interactions. Here the more intense color represents the more densely populated value for a particular class.

\subsection{General Global Rule effects}

% For two-column wide figures use
\begin{figure*}
% Use the relevant command to insert your figure file.
% For example, with the graphicx package use
  \includegraphics[width=0.75\textwidth, angle=-90]{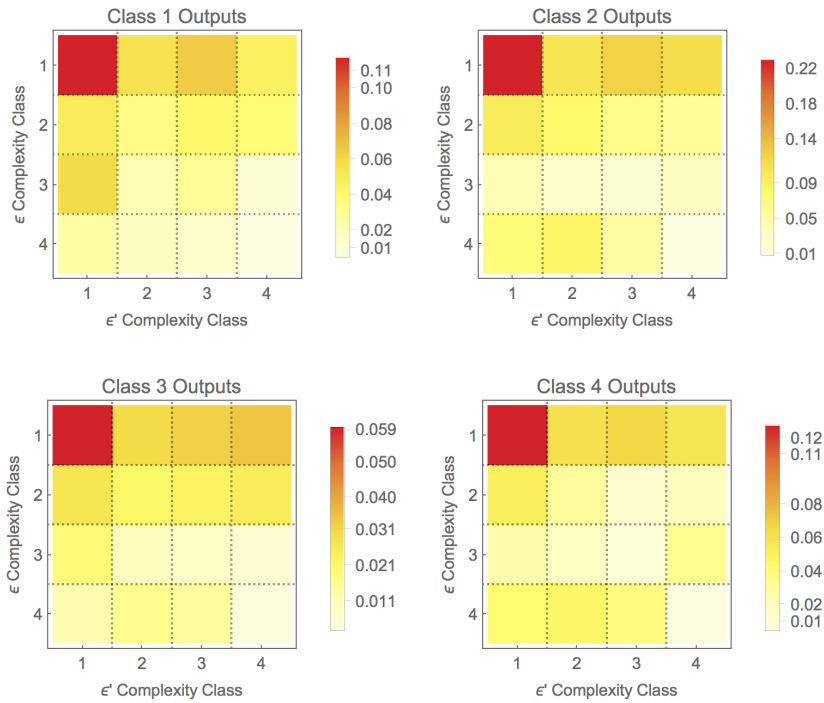}
% figure caption is below the figure
\caption{Heat maps for the 4 classes of output types for \emph{all} executions. Each heat map represents the percent of cases that resulted in the corresponding output complexity class.}
\label{fig:2}       % Give a unique label
\end{figure*}

The outputs of every possible $425\,600$ GRs are accumulated and represented in Figure 1. In general, this figure shows the change in complexity when a GR is used to determine the interaction between two ECAs. Most of the outputs for each complexity class were generated with Class 1 ECA interactions, which was least expected.

\subsection{Global Rules of interest}

% For two-column wide figures use
\begin{figure*}[h!]
% Use the relevant command to insert your figure file.
% For example, with the graphicx package use
  \includegraphics[width=0.26\textwidth, angle=-90]{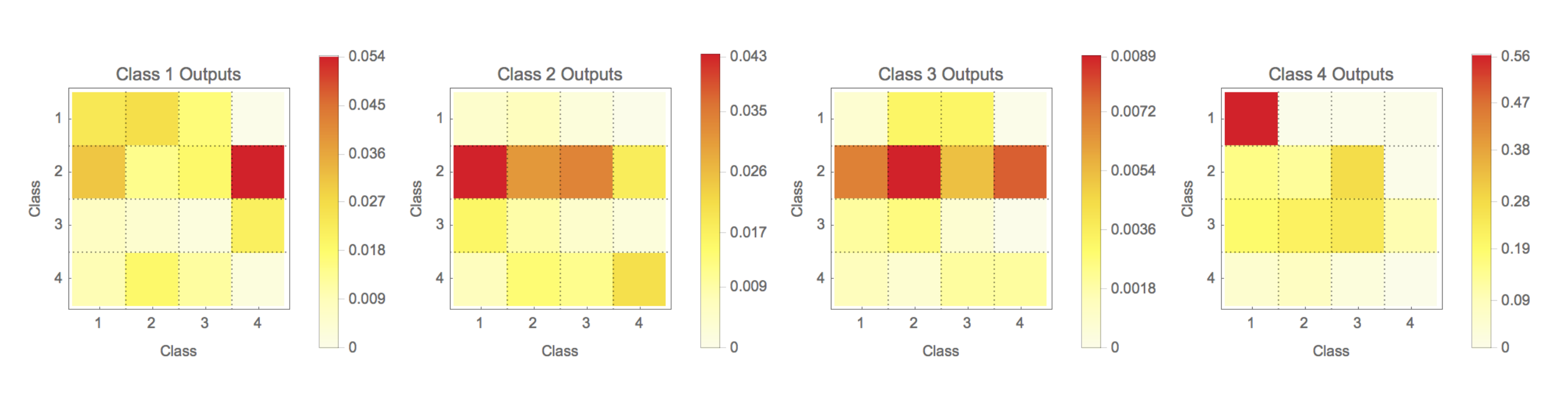}
% figure caption is below the figure
\caption{Heat maps for the 4 classes of output types for GR 36983. Each heat map represents the percent of cases that resulted in the corresponding output complexity class.}
\label{fig:3}       % Give a unique label
\end{figure*}

% For two-column wide figures use
\begin{figure*}[h!]
% Use the relevant command to insert your figure file.
% For example, with the graphicx package use
  \includegraphics[width=0.26\textwidth, angle=-90]{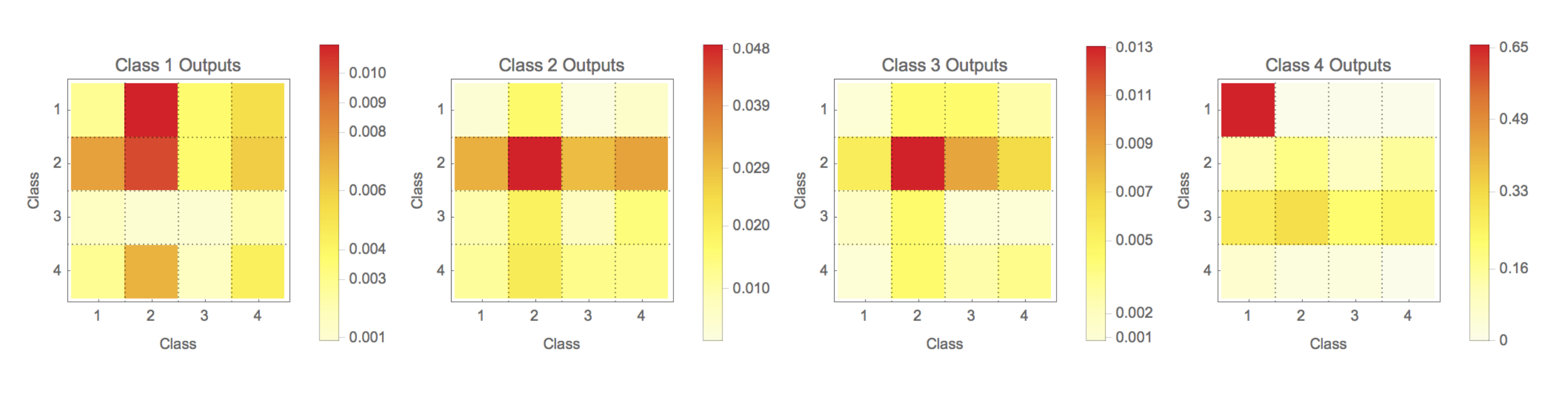}
% figure caption is below the figure
\caption{Heat maps for the 4 classes of output types for GR 72499. Each heat map represents the percent of cases that resulted in the corresponding output complexity class.}
\label{fig:4}       % Give a unique label
\end{figure*}

% For two-column wide figures use
\begin{figure*}[h!]
% Use the relevant command to insert your figure file.
% For example, with the graphicx package use
  \includegraphics[width=0.26\textwidth, angle=-90]{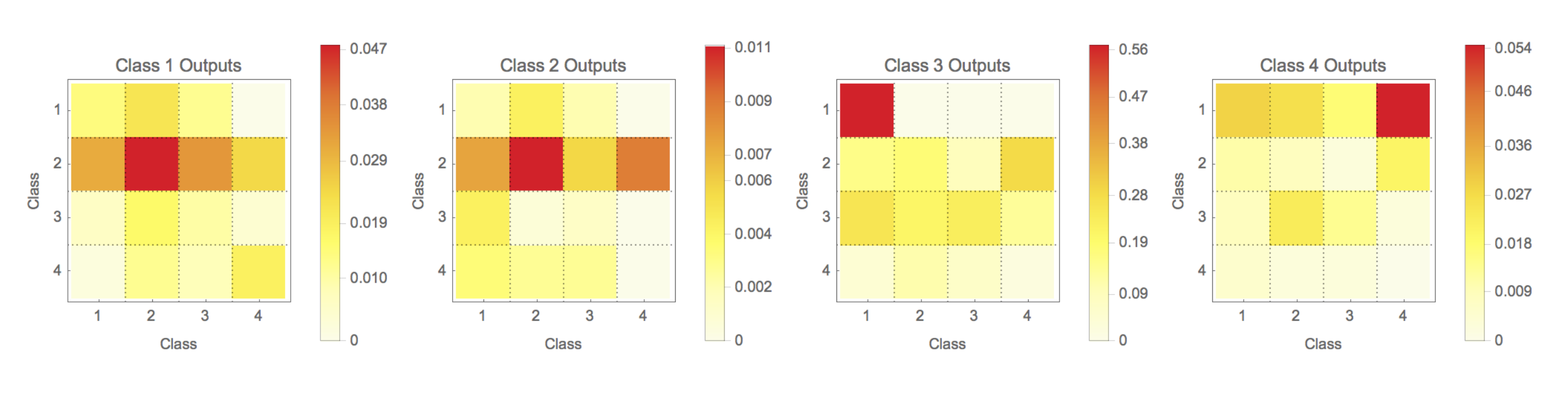}
% figure caption is below the figure
\caption{Heat maps for the 4 classes of output types for GR 77499. Each heat map represents the percent of cases that resulted in the corresponding output complexity class.}
\label{fig:5}       % Give a unique label
\end{figure*}

% For two-column wide figures use
\begin{figure*}[h!]
% Use the relevant command to insert your figure file.
% For example, with the graphicx package use
  \includegraphics[width=0.77\textwidth, angle=-90]{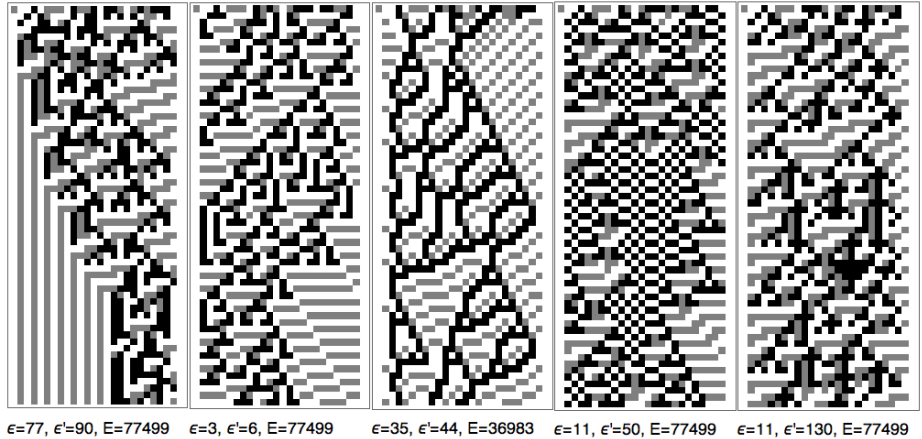}
% figure caption is below the figure
\vspace{-1.5cm}
\caption{Interesting outputs of various $\epsilon$ and $\epsilon'$ interactions and $E$. Note that most of the $\epsilon$ and $\epsilon'$ rules belong to complexity Class 1 or 2.}
\label{fig:3}       % Give a unique label
\end{figure*}

There are several interesting examples of GRs that affect the classification of behavior of the CA interaction state output. The heat maps of three GRs in particular are shown in Figures 2-4. As shown in the figure captions, GRs are enumerated according to their position in the \texttt{Tuples} function. That is to say, we enumerate the GRs by generating all tuples with 3 symbols in lexicographical order.

Over half of all outputs from with GR 77399 were complexity Class 4 from Class 1 $\epsilon$ and $\epsilon'$ interactions. The majority of complexity Class 3 outputs were generated from complexity Class 2 $\epsilon$ and $\epsilon'$ interactions. GR 72399 had outputs that were 65% Class 4 outputs, most of which also resulted from Class 1 $\epsilon$ and $\epsilon'$ interactions. GR 36983 is also similar except that over half of all outputs are complexity Class 3, which resulted from Class 1 \epsilon$ and $\epsilon'$ interactions.

The actual outputs for some of these GRs are shown in Figure 5. Note that most $\epsilon$ and $\epsilon'$ rules are complexity Class 1 or 2.

\section{Discussion}

It is unexpected that the biggest increases in complexity arise from $\epsilon$ and $\epsilon'$ complexity Class 1 interactions. This suggests that complexity is intrinsic to the GRs rather than the ECAs themselves. We suspect that interacting Class 1 ECAs readily accept complexity through the rules of their interactions. This is not as prevalent in any of the other complexity classes of $\epsilon$ and $\epsilon'$ interactions. Likely, if $\epsilon$ and $\epsilon'$ have more complexity without interaction, then they are more robust to any complexity changes introduced by the GR.

In all cases, complexity increases or remains the same by introducing an interaction rule via a GR. The most interesting cases are when Global Rules increase the complexity of the output by entire classes. There are cases where mixed neighborhoods are present and sustained throughout the output, which is a form of emergence through the interaction GR rule. Because we only used a short number of time steps per execution, it is unclear whether these mixed neighborhoods eventually die out or not, it is nonetheless a case of intermediate emergence from a Global Rule. 
 
We have found interesting cases where Global Rules seem to drastically change the complexity of an interacting CA output. Some originally Class 3 ECAs, for example, were found to be too fragile under most Global Rules, while some other are more resilient. Most importantly, the greatest increases in complexity occur when the interacting ECA are both Class 1, which is true for the majority of all possible Global Rules. Although we still have yet to understand the mechanisms behind these results, we are confident that further analysis will be important in understanding the emergence of complex structures. 

\section*{Acknowledgement}

We want to acknowledge the ASU Saguaro cluster where most of the computational work was undertaken.

\end{document}